# Stance detection in online discussions

Peter Krejzl, Barbora Hourová, Josef Steinberger

Department of Computer Science and Engineering, NTIS Center,
Faculty of Applied Sciences,
University of West Bohemia
Univerzitní 8, 306 14, Plzeň
Czech Republic

`{krejzl, steinberger}@kiv.zcu.cz`
`hourova@students.zcu.cz`

**Abstract.** This paper describes our system created to detect stance in online discussions. The goal is to identify whether the author of a comment is in favor of the given target or against. Our approach is based on a maximum entropy classifier, which uses surface-level, sentiment and domain-specific features. The system was originally developed to detect stance in English tweets. We adapted it to process Czech news commentaries.

**Paper type:** Work-in-progress paper

## 1 Introduction

Stance detection has been defined as automatically detecting whether the author of a piece of text is in favor of the given target or against it. In the third class, there are the cases, in which neither inference is likely. It can be viewed as a subtask of opinion mining and it stands next to the sentiment analysis. The significant difference is that in sentiment analysis, systems determine whether a piece of text is positive, negative, or neutral. However, in stance detection, systems are to determine author's favorability towards a given target and the target even may not be explicitly mentioned in the text. Moreover, the text may express positive opinion about an entity contained in the text, but one can also infer that the author is against the defined target (an entity or a topic). This makes the task more difficult, compared to the sentiment analysis, but it can often bring complementary information [3].

There are many applications which could benefit from the automatic stance detection, including information retrieval, textual entailment, or text summarization, in particular opinion summarization.

## 2   Task Description

The system was originally created for the SemEval 2016 task: Detecting stance in tweets [5]. The task had two independent subtasks – supervised and weakly supervised. The supervised task tested stance detection towards five targets (*Atheism, Climate Change is a Real Concern, Feminist Movement, Hillary Clinton and Legalization of Abortion)*. Participants were provided 2.814 labeled training tweets for the five targets. In the case of the weakly supervised task, there were no training data but participants could use a large number (around 70K) tweets related to the single target: *Donald Trump*. The goal was to classify tweets into three classes – *IN FAVOR, AGAINST, NONE*. The performance was measured by the average F1-score on FAVOR and AGAINST classes.

There were 19 participating systems for the supervised subtask and 9 for weakly-supervised subtask. Our system performed well for *Abortion* (2nd), *Climate change* (3rd) and *Hillary Clinton* (4th). The overall rank was 9th. In the weakly-supervised task, we were ranked 4th, only the top system was significantly better. Official results are summarized in the Table 1.

| Topic | Our system F1 (rank) | Overall F1 (rank) |
|---|---|---|
| Atheism | .5788 (8) | .6342 (9) |
| Climate change is a real concern | .4690 (3) | |
| Feminist movement | .5182 (10) | |
| Hillary Clinton | .5982 (4) | |
| Legalization of abortion | .6198 (2) | |
| Donald Trump | .4202 (4) | .4202 (4) |

**Tab 1**. Overall system performance on SemEval's Twitter data.

We used the same system to detect stance in Czech news commentaries. We collected 1.560 comments from a Czech news server[1] related to two topics – "Miloš Zeman" (the Czech president) and "Smoking ban in restaurants" (statistics in Table 2). Consider the following example from the topic "Miloš Zeman".

Target: *Miloš Zeman*
Comment: *„To je u Zemana  běžné, že používá ne pravdy! Viz Peroutka*[2]*. ..."*[3]

| TOPIC | FAVOR | AGAINST | NONE | TOTAL |
|---|---|---|---|---|
| Miloš Zeman | 180 | 170 | 300 | 750 |
| Smoking ban in restaurants | 170 | 250 | 390 | 810 |

---

[1] http://www.idnes.cz
[2] President accused famous journalist Ferdinand Peroutka (1895 - 1978) of supporting Hitler.
[3] Can be translated as: "Zeman is doing this normally – using non-truths! For example Peroutka"

**Tab 2.** Czech news commentaries data – statistics.

|  | Agreement - count | No agreement - count | Agreement [%] | Random agreement | Kappa |
|---|---|---|---|---|---|
| A2/A1 | 57 | 77 | 0.74025974 | 0.333333333 | 0.6103896 |

**Tab 3.** Agreement

## 3  The Approach Overview

We preprocessed the Czech commentaries by the same rules as in the original system [3] (for example: all urls were replaced by keyword URL, links to images are replaced by IMGURL, only letters are preserved, the rest of the characters is removed, …). Moreover, we stemmed the texts by HPS – High Precision Stemmer [2]. The system is based on a standard maximum entropy classifier [4], trained separately for each topic, with the following features.

It has been showed that **unigrams** perform quite well in this task [6]. Our model is based on TF-IDF and uses the top 1000 words from the vocabulary. The rest of the features can be turned on or off for each topic. **Initial n-grams**[4], as showed in [1] can be useful features. Out system supports initial unigrams to initial trigrams. Another surface feature was the **comment length** in words after preprocessing. We used a resource borrowed from the sentiment analysis – **Entity-centered sentiment dictionaries** (ECSD): dictionaries created mainly for the purpose of entity-related polarity detection [7].

The original system [3] used more features, which could not be easily applied on Czech commentaries. We do not work with tweets, so we could not use a set of features generated from hashtags. We have not analyzed the influence of part-of-speech (POS) tags yet. We did not identify strong candidates to build a domain specific dictionary as in [3]. Bigram features did not work in the case of the tweet analysis, so we did not use it in this work as well. However, we plan to revisit the influence of bigram, POS or domain-specific features.

## 4  Results

Table 4 shows results on the Czech data. We used two evaluation measures. The first one was used for the SemEval'16 evaluation – the average F1-score on FAVOR and AGAINST classes. The second one includes the NONE class as well. We used 10-fold cross validation to distribute training and testing data.

---
[4] Initial n-grams are basically the first n words of the sentence.

| Topic | F1 – (FAVOR/AGAINST) | F1 – (FAVOR/AGAINST/NONE) |
|---|---|---|
| Miloš Zeman | .4347 | .5204 |
| Smoking ban in restaurants | .4562 | .5400 |

**Tab 4**. System performance on Czech news commentaries.

The results show that performance on the Czech data is significantly worse (.43 – .46) than on the English tweets corpus (.47 – .62). It is mainly due to the lack of some key features like hashtags or domain-specific. Moreover, in the tweets corpus the stance tend to lean to one direction (either FAVOR or AGAINST), while in the Czech corpus most of the comments are considered neutral (NONE).

## 5    Conclusion

The paper describes the system originally created to participate in Tweet Stance Detection task in SemEval 2016 and additionally used to detect stance in Czech news commentaries. We experienced worse performance in comparison with the original English tweets corpus. It is mainly due to the lack of some significant features like hashtags. The current plan is to revisit the influence of bigram, POS or domain-specific features.

## 6    Acknowledgments

This work was supported by grant no. SGS-2013-029 Advanced computing and information systems and by project MediaGist, EU's FP7 People Programme (Marie Curie Actions), no. 630786.

*5*SemEval '16, ACL.
6. Somasundaran, S. and Wiebe, J. 2009. Recognizing stances in online debates. In Proceedings of the ACL/AFNLP, pages 226-234, ACL.
7. Steinberger, J., Lenkova, P., Ebrahim, M., Ehrmann, M., Hurriyetoglu, A., Kabadjov, M., Steinberger, R., Tanev, H., Zavarella, V. and Vazquez, S. 2012. Creating Sentiment Dictionaries via Triangulation. In Decision Support 53(4), pages 689-694, Elsevier.